\documentclass[a4paper,fleqn]{cas-dc}

\usepackage[numbers,sort&compress]{natbib}

\usepackage{xcolor,soul,framed} %,caption

\colorlet{shadecolor}{yellow}
\graphicspath{{../pdf/}{../jpeg/}}
\DeclareGraphicsExtensions{.pdf,.jpeg,.png}

\usepackage{hyperref}
\usepackage{array}
\usepackage{mdwmath}
\usepackage{mdwtab}
\usepackage{eqparbox}
\usepackage{url}
\usepackage{verbatim}
\usepackage{threeparttable}
\newcommand{\tabincell}[2]{\begin{tabular}{@{}#1@{}}#2\end{tabular}}  
\usepackage{soul}
\usepackage{multirow}
\usepackage{algorithm}
\usepackage{algpseudocode}

\def\tsc#1{\csdef{#1}{\textsc{\lowercase{#1}}\xspace}}
\tsc{WGM}
\tsc{QE}
\begin{document}

\let\WriteBookmarks\relax
\def\floatpagepagefraction{1}
\def\textpagefraction{.001}

\shorttitle{Semi-supervised landmark detection via Shape-Regulated Self-training}

\shortauthors{Chen et al.}  

\title [mode = title]{Semi-supervised Anatomical Landmark Detection via Shape-regulated Self-training}

\tnotemark[1] 

\tnotetext[1]{This work is funded by the Innovative Technology Fund of the Innovation and Technology Bureau, Hong Kong SAR.}


\author[1]{Runnan Chen}[orcid=0000-0002-8832-9016]


\ead{rnchen2@cs.hku.hk}

%\ead[url]{<URL>}

%\credit{<Credit authorship details>}

\affiliation[a]{organization={The University of Hong Kong},
            %addressline={}, 
            city={Hong Kong},
%          citysep={}, % Uncomment if no comma needed between city and postcode
            %postcode={}, 
            %state={},
            country={China}}

\author[2]{Yuexin Ma}

% Footnote of the second author

% Email id of the second author
\ead{mayuexin@shanghaitech.edu.cn}



\affiliation[b]{organization={ShanghaiTech University},
            %addressline={}, 
            city={Shanghai},
%          citysep={}, % Uncomment if no comma needed between city and postcode
            %postcode={}, 
            %state={},
            country={China}}

\author[3]{Lingjie Liu}

% Footnote of the second author

% Email id of the second author
\ead{lliu@mpi-inf.mpg.de}



\affiliation[b]{organization={Max Planck Institute for Informatics},
            %addressline={}, 
            city={Saarbrücken},
%          citysep={}, % Uncomment if no comma needed between city and postcode
            %postcode={}, 
            %state={},
            country={Germany}}

\author[1]{Nenglun Chen}

% Footnote of the second author

% Email id of the second author
\ead{chennenglun@gmail.com}



\author[1]{Zhiming Cui}

% Footnote of the second author

% Email id of the second author
\ead{cuizm.neu.edu@gmail.com}



\author[4]{Guodong Wei}

% Footnote of the second author

% Email id of the second author
\ead{g.d.wei.china@gmail.com}



\affiliation[b]{organization={ South China University of Technology},
            %addressline={}, 
            city={Guangzhou},
%          citysep={}, % Uncomment if no comma needed between city and postcode
            %postcode={}, 
            %state={},
            country={China}}

\author[1]{Wenping Wang}

% Footnote of the second author

% Email id of the second author
\ead{wenping@cs.hku.hk}
\cormark[1]



\cortext[1]{Corresponding author}

\fntext[1]{This is the first author footnote. but is common to third author as well.}
\fntext[2]{Another author footnote, this is a very long footnote and it should be a really long footnote. But this footnote is not yet sufficiently long enough to make two lines of footnote text.}

\begin{abstract}
Well-annotated medical images are costly and sometimes even impossible to acquire, hindering landmark detection accuracy to some extent. Semi-supervised learning alleviates the reliance on large-scale annotated data by exploiting the unlabeled data to understand the population structure of anatomical landmarks. The global shape constraint is the inherent property of anatomical landmarks that provides valuable guidance for more consistent pseudo labelling of the unlabeled data, which is ignored in the previously semi-supervised methods. In this paper, we propose a model-agnostic shape-regulated self-training framework for semi-supervised landmark detection by fully considering the global shape constraint. Specifically, to ensure pseudo labels are reliable and consistent, a PCA-based shape model adjusts pseudo labels and eliminate abnormal ones. A novel Region Attention loss to make the network automatically focus on the structure consistent regions around pseudo labels. Extensive experiments show that our approach outperforms other semi-supervised methods and achieves the relative improvement of 3.8\%, 6.1\% and 6.3\% on three medical image datasets. Furthermore, our framework is flexible and can be used as a plug-and-play module integrated into most supervised methods to improve performance further.
\end{abstract}


%\item 
%\item 
%\item 

% Keywords

\maketitle

%\vspace*{1ex}
\section{Introduction}

Anatomical landmarks are widely used in parametric modeling~\cite{vstern2011parametric}, segmentation~\cite{heimann2009statistical}, registration~\cite{urschler2006automatic} of medical images for bone age estimation~\cite{ebner2014towards,vstern2016automated,payer2016regressing} and quantifying various anatomical abnormalities~\cite{urschler2018integrating,ibragimov2014shape}. In practice, manually locating landmarks is a tedious and time-consuming task that requires substantial professional knowledge, resulting in unreliable quality due to the individual variation. Therefore, fully automatic and accurate detection of anatomical landmarks is an urgent and important medical image analysis task.

\begin{figure}
 %\vspace*{-5ex}
  \includegraphics[width=0.48\textwidth]{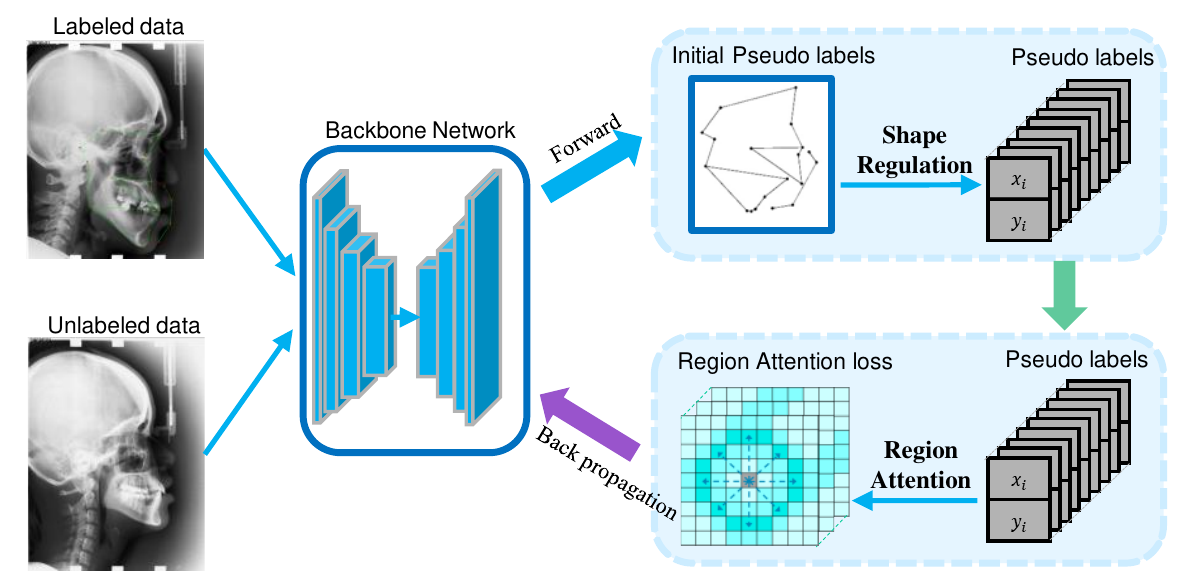}
  %\vspace*{-4ex}
  \caption{Illustration of our framework. In each iteration, the ``pseudo labels" of unlabeled data are estimated based on the backbone neural network and refined by the Shape Regulation. Then Region Attention loss is used to train the network for better predicting the unknown ground truth of unlabeled data (represented as latent variables). As the result of the iterative self-training process, the“pseudo labels” will get closer to the unknown ground truth, and we get a well-trained network at the same time.}
  \label{fig:EM_process}
  %\vspace{-2ex}
\end{figure}

Deep learning has dramatically advanced the state-of-the-art in solving anatomical landmark detection \cite{Payer2019Integrating,Chen2019cephalometric,oh2020deep,li2020structured,urschler2018integrating,newell2016stacked,chen2021structure}. They typically apply heatmaps to model the probability distribution of landmarks. A drawback of these methods is their reliance on large-scale labelled data. It is a particularly critical issue because labelling medical images are much more complicated than labelling ordinary images due to the high cost and high-level requirement of medical expertise.

Semi-supervised methods have been proposed to alleviate the reliance on labelled data by utilizing unlabeled data. The common way is to infer pseudo labels of unlabeled data for retraining the model. However, inconsistent pseudo labels would confuse and misinform the subsequent retraining procedure, resulting in even negative effort for the final performance. Some approaches utilize additional information, like the calculated optical flow in videos \cite{dong2018supervision} or extra attribute labels  \cite{honari2018improving} to enhance pseudo labelling consistency, which is usually inapplicable and unreliable for medical images. Some methods adopt co-training~\cite{dong2018few,dong2019teacher,pham2021meta} or consistency regularization~\cite{honari2018improving,radosavovic2018data,xie2020unsupervised,sohn2020fixmatch} for semi-supervised learning. However, the inherent global shape constraint of anatomical landmarks are ignored in these methods. For example, all the cephalograms have similar views, and the relative positions of the anatomical landmarks and angels of bone structure are similar for all humans (Figure \ref{fig:abnormalDetection} A). Considering those global shape constraints benefit for more reliable pseudo-labelling.

In this paper, we propose a novel shape-regulated self-training approach for semi-supervised landmark detection, which fully utilizes the global shape constraint existing in anatomical landmarks. There are two steps in the training process (Figure \ref{fig:EM_process}). In the first step, the ``pseudo labels" for unlabeled data are estimated by the pre-trained network and then regulized by a Principle Component Analysis (PCA)-based~\cite{wold1987principal} shape model. In the second step, we design a novel Region Attention loss that exploring consistent structure regions across both labelled data and pseudo-labelled data (unlabeled data with ``pseudo labels") to optimize the network parameters. As the result of iterative training, the ``pseudo labels" of unlabeled data will closer to its unknown ground truth. We obtain a well-trained network at the same time. Because extra data are rationally labelled and trained, the network outperforms the purely supervised counterpart. 

Extensive experiments conducted on 2D and 3D datasets show that our method outperforms other semi-supervised methods by large margins. Besides incorporating our method with various baselines, we achieve state-of-the-art performances on three benchmarks with the relative improvement of 3.8\%, 6.1\% and 6.3\%, respectively. Furthermore, our method performs well even with very few labelled data. For instance, it achieves 16.7\% improvement with just 20 labelled samples on the cephalogram dataset. We also show the effectiveness and generalization of our method regardless of whether the unlabeled data comes from the same device or not.

The main contributions of our work are:
\begin{itemize}
\item {We propose a novel shape-regulated self-training approach for semi-supervised landmark detection in 2D and 3D medical images and achieve state-of-the-art performance.}
\item {We introduce a new pseudo label mechanism under the constraints of the PCA-based Shape Prior.} 
\item {We propose a novel Region Attention loss for self-refinement of pseudo labels and optimizing network parameters.}
\item {Our approach is flexible and can be used as a plug-and-play module integrated into the existing heatmap-based networks for further performance improvement.}
\end{itemize}

%\vspace*{1ex}

\section{Related Work}

\subsection{Classical Method}
Traditional landmark detection methods can be classified into two categories, named model-based methods and regression-based methods.
Model-based methods such as ASM~\cite{cootes1995active}, AAM~\cite{cootes2001active},  CLM~\cite{cristinacce2006feature} 
iteratively find the best landmark positions by pre-defined templates and regulate the global spatial shape by a shape model. The regression-based methods infer landmark location from image features directly~\cite{lindner2014robust}. Although these traditional methods have achieved considerable success in landmark detection, they rely mainly on handcrafted features and have limited accuracy.

\subsection{Supervised Method}
Recent advances in deep neural networks~\cite{lecun2015deep} have achieved great success in landmark detection due to their ability to extract deep features. There are two commonly used models for landmark detection, named coordinate regression models~\cite{feng2018wing} and heat map regression models~\cite{newell2016stacked}. Coordinate regression models directly regress landmark coordinates from the input image. Compared with implicit learning landmark features in coordinate regression models, the heat map regression models can explicitly predict each landmark's probability distribution on the heat map~\cite{liu2019semantic,sun2018integral,oh2020deep}. Because landmarks on one image have a structural relationship with each other. Some methods make use of the structural constraint by incorporating structural information into deep neural networks for locating landmarks~\cite{Payer2019Integrating,li2020structured}. Chen et al.~\cite{Chen2019cephalometric} propose a framework that fuses different levels of the feature with self-attention to perform landmark detection. These fully supervised methods have produced compelling results in landmark detection. However, they require a large amount of labelled data, and marking landmarks is very costly and time-consuming, especially for medical images that require professional knowledge.

\subsection{Semi-supervised Method}
Semi-supervised learning has been explored to reduce labelling data labour and improve accuracy with additional unlabeled data in image classification~\cite{jiang2015self,kumar2010self,dong2018few,wu2017highly,berthelot2019remixmatch,sohn2020fixmatch,wu2018self,xie2020unsupervised,berthelot2019mixmatch,pham2021meta}. They typically design heuristics to select pseudo-labelled data for re-training~\cite{jiang2015self,kumar2010self,dong2018few,wu2017highly,wu2018self,pham2021meta}. Or apply consistency regularisation on the predictions with multiple data augmentation techniques. \cite{berthelot2019remixmatch,sohn2020fixmatch,xie2020unsupervised,berthelot2019mixmatch,chen2020unsupervised}. For landmark detection \cite{dong2018supervision,honari2018improving,radosavovic2018data,dong2019teacher}, some works focus on introducing additional information in a semi-supervised learning framework. Dong et al.~\cite{dong2018supervision} propose a supervision-by-registration method that uses additional unlabeled video with the optical flow as temporal consistency to enhance the accuracy of landmark identification. Honari et al.~\cite{honari2018improving} introduces a multi-tasks framework that improves the performance by extra image-level attribute labels. However, the above methods that rely on additional information, which is not available for medical images. Consistency regularization, as a kind of semi-supervised learning technique, has also been explored in landmark detection. Radosavovic et al.~\cite{radosavovic2018data} improve the quality of pseudo landmarks by ensembling predictions of multiple input data transformations. Honari et al.~\cite{honari2018improving} utilize the landmark consistency under image transformations to enlarge the dataset for improving performance. Xie et al.~\cite{xie2020unsupervised} apply data augmentation techniques on unlabeled data to improve consistency training in various deep learning tasks. Dong et al.~\cite{dong2019teacher} propose a method that teacher network filters low-confident labels for training student networks. They ensemble pseudo labels' predictions in multi-models, but we eliminate the inconsistency with a single model by shape-regulated self-training. Moreover, we deeply explored and fully used the global shape constraints of anatomical landmarks in our method.

\section{Method}
Given some labeled data $S_{l}=\{I^{l}_{1}, I^{l}_{2}, ..., I^{l}_{N}\}$ and unlabeled data $S_{u}=\{I^{u}_{1}, I^{u}_{2}, ..., I^{u}_{M}\}$, where $N$ and $M$ are the number of samples. We aim at training a neural network $\mathcal{F}$ to minimize the prediction errors for both $S_{l}$ and $S_{u}$. The objective function of the optimal network parameters $\theta^{*}$ is:
\begin{equation}\label{equ:objective_Function}
\begin{aligned}
\theta^{*}=&\mathop{\arg\min}_{\theta}\frac{1}{N}\sum_{i=1}^{N}|\mathcal{F}(I^{l}_{i};\theta) - G_{i}| + \frac{1}{M}\sum_{j=1}^{M}|\mathcal{F}(I^{u}_{j};\theta) - Q_{j}|,
\end{aligned}
\end{equation}
where $G_{i}=\{g^{i}_{1},g^{i}_{2},...,g^{i}_{n}\}$ are the ground truth landmarks of the labeled sample $I^{l}_{i}$, and $n$ is the number of landmarks. $Q_{j}=\{q^{j}_{1},q^{j}_{2},...,q^{j}_{n}\}$ are unknown ground truth for unlabeled sample $I^{u}_{j}$. 

Network $\mathcal{F}$ predicts $n$ heatmaps for $n$ landmarks. And each heatmap reflects the probability distribution of a landmark. We transfer the $t$-th heatmap $\mathcal{H}_{t}$ to the $t$-th landmark coordinate by the integral operation:

\begin{equation}\label{equ:integralFunction}
\mathcal{F}(I;\theta) = \frac{1}{\gamma_t}\sum_{\rho\in \mathcal{H}_{t}}\mathcal{H}_{t}(\rho|I;\theta)*\rho, \gamma_t = \sum_{\rho\in \mathcal{H}_{t}}\mathcal{H}_{t}(\rho|I;\theta),
\end{equation}
where $\mathcal{H}_{t}(\rho|I;\theta)$ is the value of the predicted heatmap $\mathcal{H}_{t}$ at the pixel $\rho$, and $\gamma_t$ is the spacial normalization term.

As shown in the objective function (\ref{equ:objective_Function}), the neural network $\mathcal{F}$ with $\theta^{*}$ will be more powerful as more data are involved in training. However, $Q_{j}$ are unknown actually. To solve this, our framework takes $Q_{j}$ as latent variables, which are represented as the sum of ``pseudo labels" and pre-defined offsets. As the result of the iterative self-training process, the ``pseudo labels" will get closer to the unknown ground truth. And we get a well-trained network with $\theta^{*}$ at the same time. 

In the following sections, we first present how to obtain reliable pseudo labels of unlabeled data. Then we describe how to optimize the network by Region Attention loss. Finally, we introduce the training protocol for our framework.

\subsection{Shape Regulation for Reliable Pseudo Labeling}

\begin{figure}
  %\vspace*{-5ex}
  \includegraphics[width=0.48\textwidth]{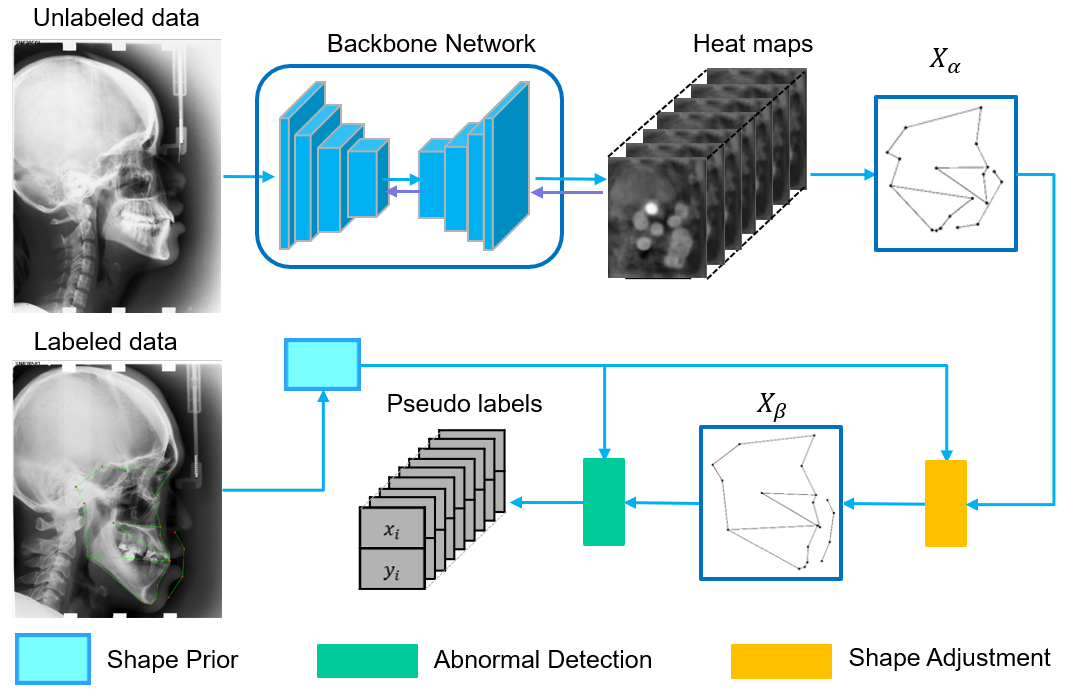}
  %\vspace*{-4ex}
  \caption{Illustration of the shape regulated pseudo labelling. The pre-trained backbone network infers the initial pseudo labels of unlabeled data. More reliable pseudo labels are obtained by shape adjustment and abnormal detection under our proposed Shape Prior. }
  \label{fig:EstimationStep}

\end{figure}

The Shape Prior is used to obtain reliable pseudo labels of unlabeled data. It filters out the low-quality pseudo labels by two steps: i) reasonable shape adjustment based on global shape structure, and ii) abnormal detection based on the deviation of individual pseudo labels. We present the details of Shape Prior, shape adjustment and abnormal detection in the following. (Figure~\ref{fig:EstimationStep}).

\subsubsection{\textbf{Shape Prior}}
Images in landmark detection tasks have inherent structural information. For example, all the cephalograms have similar views, and the relative positions of the eyes, nose, mouth on the face are similar for all humans. Suppose we have a set of cephalograms with landmark labels and a set of cephalograms without labels. The shape model behind landmarks of labelled data will have effective guidance for labelling unlabeled data. Based on this observation, we construct a shape model from labelled data and take the shape model as the Shape Prior to regulate initial pseudo labels $x_\alpha$ and make $x_\alpha$ in accord with the shape rules more. 

Considering that there is insufficient labelled data to train a deep learning-based shape model, we construct a PCA-based shape model to regulate the initial pseudo labels $X_\alpha$ of unlabeled data. The PCA-based shape model built on $S_{l}$ is denoted as $X=\{x_{1}, y_{1}, x_{2}, y_{2},..., x_{n}, y_{n}\}$, where $n$ is the number of landmarks. We can encode the shape model as follows:
\begin{equation}\label{equ:shapedefine}
X=T_{\theta}(\overline{X} + Pb)
\end{equation}
where $\overline{X}$ is the mean shape of all labeled data, $P$ is the first $K$ principal components that contain 99.99\% of variation, $T_{\theta}$ is an affine transformation (e.g. scaling, translation, rotation), $b=\{e_{1}, e_{2},...,e_{K}\}$ is the shape model parameter. Different values of $b$ will cause different shapes varying from the mean shape. It can well describe the similarity and difference of shapes belonging to the same class.

\subsubsection{\textbf{Shape Adjustment}}

The initial pseudo labels $X_{\alpha}$ of an unlabeled sample is predicted by the pre-trained network (formula \ref{equ:integralFunction}). $b_{\alpha}$ is obtained by:
%\vspace{-0.5ex}
\begin{equation}\label{equ:calculateB}
b_{\alpha} = \mathcal{P}^{-1}(T_{\alpha}X_{\alpha} - \overline{X}),
\end{equation}
where $T_{\alpha}$ is the affine transformation from $X_{\alpha}$ to $\overline{X}$, which is calculated by Procrustes analysis \cite{gower1975generalized}. 

We regulate the shape parameter $b_{\alpha} = \{e_{1}^{\alpha}, e_{2}^{\alpha},...,e_{K}^{\alpha}\}$ to perform shape adjustment on $X_{\alpha}$. Assume that the $e_k, k=1,2,...,K$ of all labeled samples ($N$ samples) is denoted as $E_k = \{e_{k}^1, e_{k}^2,...,e_{k}^N\}$. We find that $E_k$ follows the Gaussian distributions by Shapiro-Wilk test~\cite{razali2011power}, e.g. $e_{k} \sim \mathcal N (0,\sigma_k)$, where $\sigma_{k}$ is the standard deviation of $E_{k}$. Thus the majority of (99.73\%) shape parameters of $E_{k}$ is supposed to be in the range of $(-3\sigma_{k}, +3\sigma_{k})$ according to the 3$\sigma$ principle. Therefore, each component of $b_{\alpha}$ is adjusted to its corresponding 3$\sigma$ range if it does not locate in the range to form a more reasonable shape parameter $b_{\beta}=\{e_{1}^{\beta}, e_{2}^{\beta},...,e_{K}^{\beta}\}$. 

\begin{equation}\label{equ:adjustE}
e_{k}^{\beta}=\left\{
\begin{array}{rcl}
-3\sigma_{k} & & {e_{k}^{\alpha} \leq -3\sigma_{k}}\\
e_{k}^{\alpha} & & {-3\sigma_{k}< e_{k}^{\alpha} < +3\sigma_{k}}\\
+3\sigma_{k} & & {+3\sigma_{k} \leq e_{k}^{\alpha}}
\end{array} \right.
\end{equation}

The adjusted shape $X_{\beta}$ is calculated by following:
%\vspace{-1ex}
\begin{equation}\label{equ:calculateShapeBeta}
X_{\beta}=T_{\alpha}^{-1}(\overline{X} + \mathcal{P}b_{\beta}).
\end{equation}

\begin{figure}
 %\vspace*{-5ex}
  \includegraphics[width=0.48\textwidth]{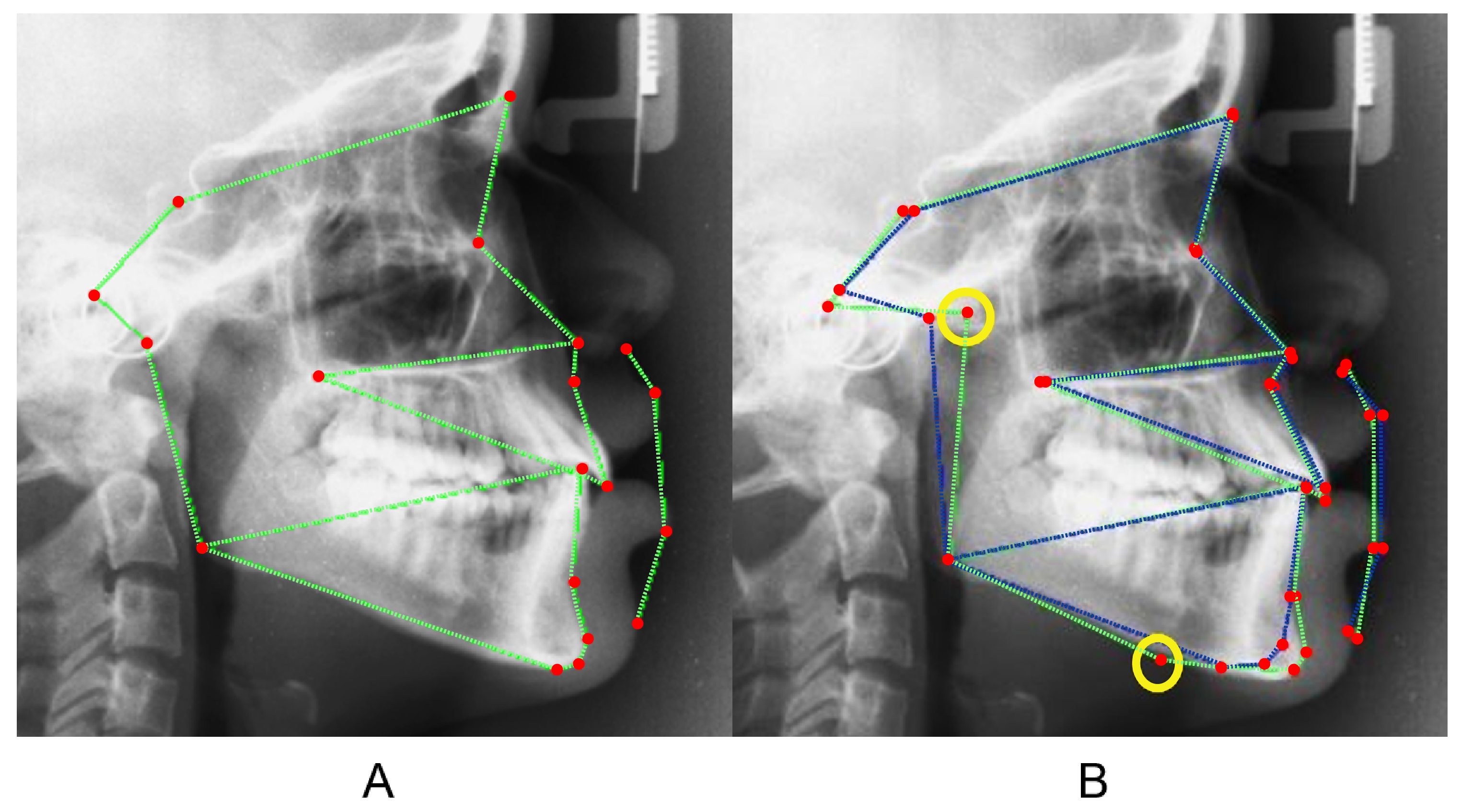}
  \vspace*{-3ex}
  \caption{Illustration of Shape Prior regulation. Red points in Sub-figure A are the ground truth labels. The green shape and blue shape in Sub-figure B present the initial pseudo labels $X_{\alpha}$ and the adjusted pseudo labels $X_{\beta}$ respectively. Abnormal pseudo labels with relatively large offsets are in yellow circles.}
  \label{fig:abnormalDetection}
  \vspace*{-3ex}
\end{figure}

\subsubsection{\textbf{Abnormal Detection}}

We observe that if the shape $X_\alpha$ fits the shape prior well, all the landmarks' deviation would be small after the adjustment. Otherwise, one or more landmarks will deviate from the original positions too much. And the adjustment of such landmarks will wrongly affect the adjustment of other landmarks, which causes an invalid $X_\beta$ (Figure~\ref{fig:abnormalDetection} B). Therefore, we propose the following heuristics to get the final pseudo labels $X_{\gamma}$ of $S_{u}$. \

\begin{itemize}

\item 
$X_{\gamma} = X_{\beta}$, if the deviation of all the landmarks are no larger than $Z$.
\item 
$X_{\gamma} = X_{\alpha}$ excludes those landmarks whose deviation is larger than $Z$, if the deviation of any landmark is larger than $Z$.

\end{itemize}
The threshold $Z$ is set to be 2mm error as it is clinically acceptable precision in a medical image. Then, we get the final reliable pseudo labels $X_{\gamma}$:
\begin{equation}\label{equ:calculateXGamma}
X_{\gamma}=J(\mathcal{F}(I^{u};\theta)),
\end{equation}
where $J(\bullet)$ stands for the joint operations of Shape Adjustment and Abnormal Detection, $I^{u}$ is the unlabeled sample, and $\mathcal{F}$ is the network with parameters $\theta$.

\subsection{Self-training via Region Attention Loss} 

\subsubsection{\textbf{Latent Variables}}

Although reliable pseudo labels $X_{\gamma}$ are obtained, there are still offsets between $X_{\gamma}$ and the unknown ground truth $Q$. We assume that the offsets follow pre-defined Gaussian Distributions. The latent variables $Q$ are formulated as follows:
\begin{equation}\label{equ:latentVariables}
Q=X_{\gamma} + \Delta,
\end{equation}
where $\Delta=\{\delta_{1}, \delta_{2},...,\delta_{n}\}$, and $|\delta_{i}| \sim N(0.01, 0.005^2) $, $i=1,2,...,n$. The mean 0.01 is determined empirically.

\begin{figure}
  %\vspace*{-5ex}
  \includegraphics[width=0.48\textwidth]{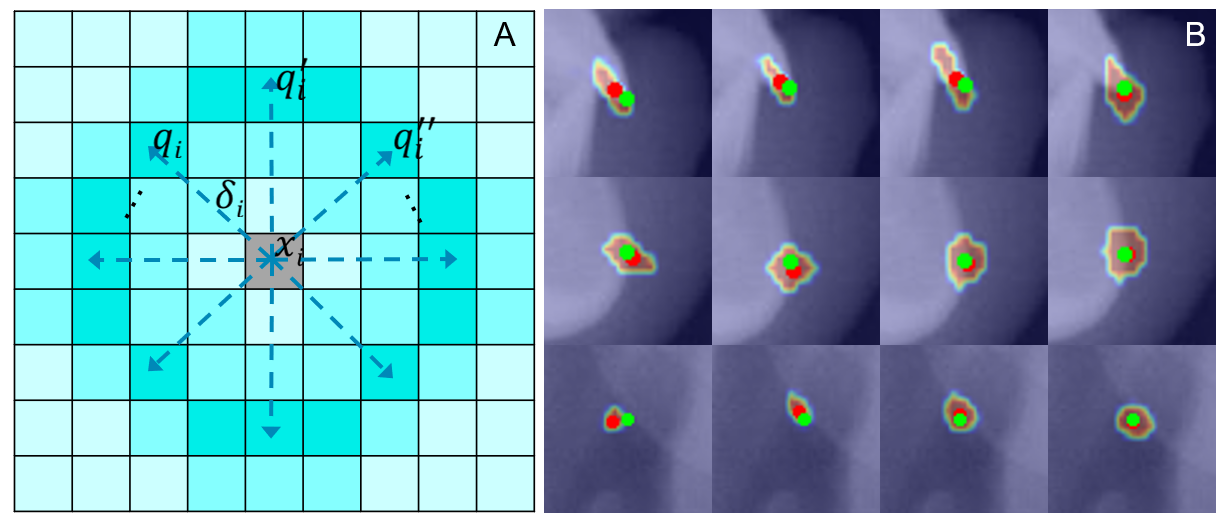}
  \vspace*{-3ex}
  \caption{Illustration of our Region Attention loss. In sub-figure A, $x_{i}$ is donated as the pseudo label, and $q_{i}, q_{i}^{'}, q_{i}^{''}$ et al. represent multiple latent unknown ground truth. In sub-figure B, each row presents one landmark prediction of an unlabeled sample during the self-training process  (from the left column to the right column are the results of 5, 10, 20, 70 epochs, respectively). The red points are the predicted pseudo labels of unlabeled data, while the green points are the human annotations (they are not used for training, just for reference). It shows that the predicted pseudo labels are self-refined and gradually approach the unknown ground truth during the self-training process.}
  \label{fig:RegionAttention}
  \vspace*{-3ex}
\end{figure}

\begin{algorithm*}
\caption{Shape-regulated Self-training}\label{alg:SelfTraining}
{\bf Input:}
Batch of labeled examples and their labels \{($I^{l}_{i}$, $G_{i}$), i=1, 2, 3,...,N\}, batch of unlabeled examples \{$I^{u}_{j}$, j=1, 2, 3,...,M\}, the network $\mathcal{F}$ with parameters $\theta$.

{\bf Initialization:} \\
$\theta=\mathop{\arg\min}_{\theta}\frac{1}{N}\sum_{i=1}^{N}|\mathcal{F}(I^{l}_{i};\theta) - G_{i}|$ \Comment{Pre-training} \\
$X=T_{\theta}(\overline{X} + Pb)$ \Comment{PCA Shape Prior modeling} \\
{\bf Self-training:}

\begin{algorithmic}
\Repeat
\For{$j = 1$ do $M$} \Comment{Sample unlabeled data $I^{u}_{j}$ with data augmentation} \\ 

\State $X_{\alpha} \gets \frac{1}{\gamma_t}\sum_{\rho\in \mathcal{H}_{t}}\mathcal{H}_{t}(\rho|I^{u}_{j};\theta)*\rho,    \gamma_t = \sum_{\rho\in \mathcal{H}_{t}}\mathcal{H}_{t}(\rho|I^{u}_{j};\theta)$ \Comment{Initial Pseudo-labeling}

\State $X_{\beta} \gets T_{\alpha}^{-1}(\overline{X} + \mathcal{P}b_{\beta}), b_{\alpha} = \mathcal{P}^{-1}(T_{\alpha}X_{\alpha} - \overline{X}), b_{\beta}=\{e_{1}^{\beta}, e_{2}^{\beta},...,e_{K}^{\beta}\}$ \Comment{Shape adjustment}

\If{if the deviation of all the landmarks are no larger than $Z$}  \Comment{Abnormal landmark filtering}
    \State $X_{\gamma} \gets X_{\beta}$
\ElsIf{if the deviation of any landmark is larger than $Z$}
    \State $X_{\gamma} \gets X_{\alpha}$ excludes those landmarks whose deviation is larger than $Z$
\EndIf
\State $Q_{j} \gets X_{\gamma} + \Delta$ \Comment{Latent variables}
\EndFor
\State $\theta \gets \mathop{\arg\min}_{\theta}\frac{1}{N}\sum_{i=1}^{N}|\mathcal{F}(I^{l}_{i};\theta) - G_{i}| + \frac{1}{M}\sum_{j=1}^{M}|\mathcal{F}(I^{u}_{j};\theta) - Q_{j}|$ \Comment{Optimisation by region attention loss}
\Until{Convergence}
\end{algorithmic}
{\bf Fine-tuning:} \\
$\theta=\mathop{\arg\min}_{\theta}\frac{1}{N}\sum_{i=1}^{N}|\mathcal{F}(I^{l}_{i};\theta) - G_{i}|$ \\
{\bf Output:} $\theta$ \Comment{The final network parameters}
\end{algorithm*}

\subsubsection{\textbf{Region Attention Loss}} 
We propose a Region Attention loss to effectively train the pseudo-labeled data. Given the sample $I$, we estimate pseudo labels $X_{\gamma}=\{x^{\gamma}_{1}, x^{\gamma}_{2},...,x^{\gamma}_{n}\}$ by formula~\ref{equ:calculateXGamma} and its unknown ground truth $Q=\{x^{\gamma}_{1}+\delta_{1},x^{\gamma}_{2}+\delta_{2},...,x^{\gamma}_{n}+\delta_{n} \}$ by formula~\ref{equ:latentVariables}. To minimize the prediction errors, the expectation of network parameter $\theta^{*}$ is:
%\vspace*{0.5ex}
\begin{equation}\label{equ:expectedTheta}
\begin{aligned}
&\theta^{*} =\mathop{\arg\min}_{\theta}|F(I;\theta) - Q|=\mathop{\arg\min}_{\theta}| Q-F(I;\theta) |\\
&\Leftrightarrow \mathop{\arg\min}_{\theta} \sum_{i=1}^{n} |q_{i}*\sum_{\rho\in H_{i}}\frac{H_{i}(\rho|I;\theta)}{\gamma_i}-\sum_{\rho\in H_{i}}\frac{H_{i}(\rho|I;\theta)}{\gamma_i}*\rho| \\
&\Leftrightarrow \mathop{\arg\min}_{\theta}\sum_{i=1}^{n} ||\delta_{i}|-\sum_{\rho\in H_{i}}\frac{H_{i}(\rho|I;\theta)}{\gamma_i}*|\rho-x_{i} ||
\end{aligned}
\end{equation}

Therefore, our Region Attention loss is defined as follows:
\begin{equation}\label{equ:RegionAttentionloss}
\mathcal{L}_{R} =\sum_{i=1}^{n} ||\delta_{i}|-\sum_{\rho\in \mathcal{H}_{i}}\frac{\mathcal{H}_{i}(\rho|I;\theta)}{\gamma_i}*|\rho-x^{\gamma}_{i} ||.
\end{equation}
%\vspace*{0.5ex}

Essentially, the Region Attention loss is a soft constraint that encourages an activated region on heatmap $\mathcal{H}_{i}$ near the pseudo label $x^{\gamma}_{i}$ but is not necessarily the fixed distribution centred on the pseudo label. We find the property helps the network to focus on the structure consistent region across the unlabeled data (Figure~\ref{fig:RegionAttention}). 

In each iteration of self-training, the unknown ground truth is estimated based on the current neural network and Shape Regulation. Then Region Attention loss is used to optimize the network parameters for better predicting the unknown ground truth. As a result, the pseudo labels are self-refinement and gradually closed to the unknown ground truth though iteratively self-training (Figure~\ref{fig:RegionAttention} B). We could not get such good performance when replacing it with some hard constraint (L1, BCE). The Region Attention loss is a vital component in our framework, and its effectiveness is supported by ablation studies (`Ablation study' in the `Experiments' section).

\subsection{Training Protocol}
Our framework is trained by three sequential stages: pre-training stage, self-training stage, and fine-tuning stage (Algorithm \ref{alg:SelfTraining}). The backbone network $F$ is shared across three stages and optimized by Adam optimizer \cite{kingma2014adam} with the default configuration on the PyTorch platform. As the self-training process has been described above, we introduce the backbone network, pre-training, and fine-tuning.

\subsubsection{\textbf{Backbone network}} The input to the backbone network is a medical image, like a CT scan or an x-ray image. The outputs are $n$ heat maps with the same size as the input. Each pixel value $\mathcal{H}_{i}(\rho)$ of the heat map $\mathcal{H}_{i}$ is scaled to $(0, 1)$ by the sigmoid activation function, indicating the probability distribution for $i$-th landmark. The backbone network can be classical heat map-based neural networks, such as U-Net \cite{ronneberger2015u}, Hourglass \cite{newell2016stacked}, or other state-of-the-art network structures of landmark detection on medical images, e.g., AFPF-RV \cite{Chen2019cephalometric} and SCN \cite{Payer2019Integrating}. 
Experiments show that our approach has great flexibility as it is well performed on different backbone networks (the Experiments section). 

\subsubsection{\textbf{Pre-training and Fine-tuning}} The network is pre-trained on labelled data $S_{l}$ by L1 loss function or the original loss function of AFPF-RV and SCN. After the self-training process, we fine-tune the network on $S_{l}$ with Region Attention loss. We train the backbone network until it converges in all stages (200 epochs in each stage). Since more data are involved in the self-training process, which exploits the unlabeled data to gain more understanding of the population structure of landmarks, the best performance in the fine-tuning stage will be better than that in the pre-training stage (the Experiment section).

\section{Experiments}
Our method is integrated into various backbone networks tested on three datasets to compare with other semi-supervised methods (Table~\ref{tab:tab1}). We also demonstrate performance improvement with different amounts of labelled data (Table~\ref{tab:tab4}). The contribution of each module is verified in ablation studies (Figure~\ref{fig:ablation}). Furthermore, we also show the effectiveness regardless of whether the unlabeled data comes from the same device or not (Cephalogram and Hand X-ray dataset in Table~\ref{tab:tab1}). By incorporating our method with various baselines, we achieve state-of-the-art performances on all three benchmarks (Table~\ref{tab:tab1}). The details of the experimental setting are presented in the following sections.

\subsection{Datasets}
\begin{figure}
 %%\vspace*{-5ex}
  \includegraphics[width=0.48\textwidth]{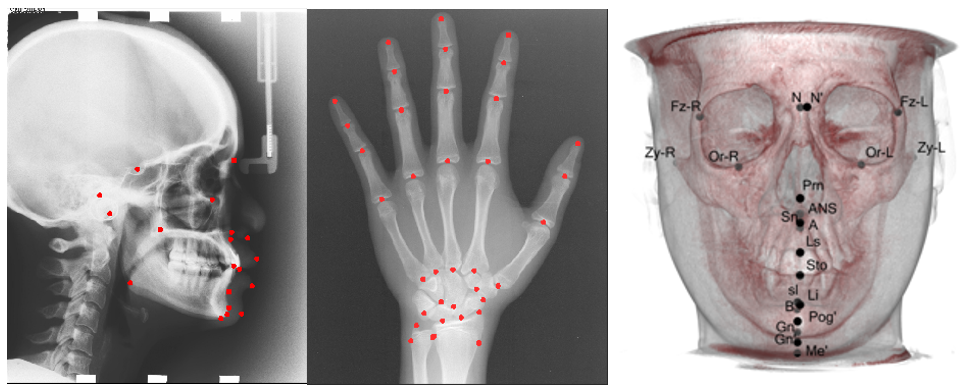}
  %%\vspace*{-4ex}
  \caption{Data samples of three datasets we used. i.e. 2D Cephalogram dataset(left), Hand X-ray dataset(middle) and 3D Head CBCT dataset(right).}
  \label{fig:datasample}
  \vspace*{-2ex}
\end{figure}

Our method is evaluated on two public 2D datasets and one in-house 3D dataset, including Cephalogram dataset \cite{lindner2016fully} , Hand X-ray dataset \cite{urschler2018integrating}, and Head CBCT dataset (Figure~\ref{fig:datasample}).

\subsubsection{\textbf{Cephalogram dataset}} It is a public dataset for cephalometric landmark detection provided by IEEE ISBI 2015 Challenge \cite{wang2016benchmark}. It contains 400 cephalometric radiographs, each with a resolution of 1935 $\times$ 2400 and 19 annotated landmarks. The spacing is 0.1mm, and the ground truth is the average annotation by two doctors. Following Payer et al. \cite{Payer2019Integrating}, we divide the dataset into two parts: 150 images for training and 250 images for testing. To assess our method with unlabeled data from other devices, we use 150 in-house cephalograms as unlabeled data.

\subsubsection{\textbf{Hand X-ray dataset}} It is a public dataset for hand anatomical landmark localization. The dataset consists of 1385 left-hand radiographs with an average resolution of 1563 $\times$ 2169, acquiring from different X-ray scanners. Among them, there are 490 unlabeled samples and 895 labelled samples. Each labelled sample has 37 manually annotated landmarks on bone joints and fingertips for the patient's age estimation. Following Payer et al.\cite{Payer2019Integrating} we use 895 labelled samples to perform three cross-validations and use the remaining 490 unlabeled samples for the self-training process in each validation.

\begin{table}[h]
    %\scriptsize
	\centering
	\caption{Comparison with state-of-the-art semi-supervised methods on Cephalogram dataset, Hand X-ray dataset, and Head CBCT dataset. $\ddagger$ indicates semi-supervised methods. Note that we show the percentage of the number of predictions larger than radius error $r$(mm) in the Cephalogram dataset.}\label{tab:tab1}
	\scalebox{0.77}{
	\begin{tabular}{l c c c c c}
		\hline
		%\multirow{2}*{\tabincell{c}{Dataset}}
		
		\multicolumn{6}{c}{\multirow{2}*{Cephalogram dataset}}\\
		~ & ~ & ~ & ~ & ~ & ~ \\
		%\multicolumn{6}{c}{\multirow{2}*{Test dataset 2}}
		
		%\multirow{2}*{\tabincell{c}{Dataset}} &
		%\multirow{2}*{\tabincell{c}{MRE +- SD}} & \multicolumn{4}{c}{$\O_{r}$ in \%} \\
        %\cline{3-6}
		%\multirow{2}*{\tabincell{c}{VGG19\_bn}} &  \multicolumn{10}{c|}{~}\\
		%~ & ~ & 2mm & 2.5mm & 3mm & 4mm\\
		
		\hline
		\multirow{2}*{\tabincell{c}{Method}} &
		\multirow{2}*{\tabincell{c}{MRE(SD)}} & \multicolumn{4}{c}{$\O_{r}$ in \%} \\
        \cline{3-6}
		%\multirow{2}*{\tabincell{c}{VGG19\_bn}} &  \multicolumn{10}{c|}{~}\\
		~ & ~ & 2mm & 2.5mm & 3mm & 4mm\\
		\cline{1-6}
		Ibragimov et al. \cite{ibragimov2014automatic}& - & 31.87 & 25.37& 20.23& 13.13\\
		Lindner et al. \cite{lindner2014robust}& - & 29.35 & 23.07& 17.83& 10.15\\
		Urschler et al. \cite{urschler2018integrating} & - & 29.79 & 23.05& 17.92& 10.99\\
		Payer et al. \cite{Payer2019Integrating} & - & 26.67 & 21.24& 16.76& 10.25\\
		Chen et al. \cite{Chen2019cephalometric} & 1.29(1.02) & 17.97 & 11.26& 7.26& 2.86\\
		Oh et al. \cite{oh2020deep} & 1.28(-) & 17.92 & 11.94& 7.66& 3.08\\
		Li et al. \cite{li2020structured} & 1.20(-) & 16.27 & 10.66& 7.28& 3.22\\

		\cline{1-6}
		U-Net \cite{ronneberger2015u}& 1.44(1.18) & 22.51 &15.58 &10.87 & 4.97\\
		U-Net+UDA$\ddagger$ \cite{xie2020unsupervised} & 1.42(1.17) & 21.98 & 15.23& 10.56& 4.77\\
		U-Net+TS$\ddagger$ \cite{dong2019teacher} & 1.41(1.15) & 22.08 & 15.11& 10.51& 4.72\\
		U-Net+Fixmatch$\ddagger$ \cite{sohn2020fixmatch} & 1.40(1.13) & 21.83 & 15.08& 10.34& 4.66\\
		U-Net+Meta$\ddagger$ \cite{pham2021meta} & 1.38(1.12) & 21.67 & 15.03& 10.15& 4.65\\
		
		U-Net+ours$\ddagger$ & 1.35(1.12) & 21.29 & 14.81& 10.08& 4.51\\
		\cline{1-6}
		AFPF-RV \cite{Chen2019cephalometric} &1.27(0.99)&18.34&11.37&7.35&2.76\\
		AFPF-RV+UDA$\ddagger$ \cite{xie2020unsupervised}& 1.25(0.98)&18.06 &11.18 &7.11 &2.61\\
		AFPF-RV+TS$\ddagger$ \cite{dong2019teacher}& 1.24(1.03)&17.97 &11.23 &7.25 &2.64\\
		AFPF-RV+Fixmatch$\ddagger$ \cite{sohn2020fixmatch}& 1.23(1.03)&17.84 &11.09 &7.18 &2.67\\
		AFPF-RV+Meta$\ddagger$ \cite{pham2021meta}& 1.21(1.01)&17.58 &11.05 &7.13 &2.56\\
		AFPF-RV+ours$\ddagger$& \textbf{1.19(0.91)}&\textbf{17.23} &\textbf{10.74} &\textbf{6.78} &\textbf{2.46}
		
	\end{tabular}}

	\scalebox{0.76}{
	\begin{tabular}{l c c c c}
		\hline
		%\multicolumn{5}{c}{}\\
		\multicolumn{5}{c}{\multirow{2}*{Hand X-ray dataset}}\\
		%\multirow{2}*{\multicolumn{5}{c}{Hand X-ray dataset}}
		~ & ~ & ~ & ~ & ~ \\
		\hline
		\multirow{2}*{\tabincell{c}{Method}} &
		\multirow{2}*{\tabincell{c}{MRE(SD)}} & \multicolumn{3}{c}{$\O_{r}$(in \%)} \\
        \cline{3-5}
		%\multirow{2}*{\tabincell{c}{VGG19\_bn}} &  \multicolumn{10}{c|}{~}\\
		~ & ~ & 2mm & 4mm & 10mm\\
		\cline{1-5}
		Payer et al. \cite{payer2016regressing}&1.13(0.98)  &4109(12.4)&444(1.34)&12(0.04)\\
		Ebner et al. \cite{ebner2014towards} & 0.97(2.45) & 2781(8.40)& 716(2.16)& 228(0.69)\\
		Stern et al. \cite{vstern2016local} &0.80(0.91) & 2582(7.80)& 512(1.55)& 15(0.05)\\
		Urschler et al. \cite{urschler2018integrating}& 0.80(0.93)&2586(7.81) &510(1.54) &18(0.05)\\
		Lindner et al. \cite{lindner2014robust}& 0.85(1.01) & 2094(6.32) & 347(1.05)& 20 (0.06)\\
		Payer et al. \cite{Payer2019Integrating} & 0.66(0.74)&1659(5.01) &241(0.73) &3(0.01)\\
		
		\hline
		Hourglass \cite{newell2016stacked} &0.75(0.56) & 1685(5.09)& 159(0.48)& 10(0.03)\\
		Hourglass+UDA$\ddagger$ \cite{xie2020unsupervised} & 0.74(0.55)&1645(4.97)&148(0.45)&8(0.03)\\
		Hourglass+TS$\ddagger$ \cite{dong2019teacher} & 0.74(0.53)&1632(4.93)&142(0.43)&9(0.03)\\
        
        Hourglass+Fixmatch$\ddagger$ \cite{sohn2020fixmatch} & 0.74(0.55)&1640(4.94)&139(0.42)&8(0.03)\\
		Hourglass+Meta$\ddagger$ \cite{pham2021meta} & 0.72(0.55)&1600(4.83)&123(0.37)&7(0.03)\\
		
		Hourglass+ours$\ddagger$ & 0.71(0.55)&1540(4.65)&93(0.28)&5(0.02)\\
		\hline
		U-Net& 0.66(0.50) & 1335(4.03) &156(0.47)& 6(0.02)\\
		U-Net+UDA$\ddagger$ \cite{xie2020unsupervised} & 0.65(0.49)&1314(3.97)&148(0.45)&8(0.02)\\
		U-Net+TS$\ddagger$ \cite{dong2019teacher} & 0.65(0.51)&1294(3.91)&135(0.41)&7(0.02)\\
		
		U-Net+Fixmatch$\ddagger$ \cite{sohn2020fixmatch} & 0.64(0.46)&1292(3.90)&142(0.43)&8(0.02)\\
		U-Net+Meta$\ddagger$ \cite{pham2021meta} & 0.64(0.45)&1271(3.84)&135(0.41)&7(0.02)\\
		
		U-Net+ours$\ddagger$ & \textbf{0.62(0.42)}& \textbf{1222(3.69)}& \textbf{102(0.31)}& \textbf{1(0.01)}\\

	\end{tabular}}
	\scalebox{0.76}{
	\begin{tabular}{l c c c c}
		\hline
		\multicolumn{5}{c}{\multirow{2}*{Head CBCT dataset}}\\
		~ & ~ & ~ & ~ & ~ \\
		\hline
		\multirow{2}*{\tabincell{c}{Method}} &
		\multirow{2}*{\tabincell{c}{MRE(SD)}} & \multicolumn{3}{c}{$\O_{r}$(in \%)} \\
        \cline{3-5}
		%\multirow{2}*{\tabincell{c}{VGG19\_bn}} &  \multicolumn{10}{c|}{~}\\
		~ & ~ & 2mm & 4mm & 8mm\\
		\cline{1-5}
		3D U-Net \cite{cciccek20163d}& 2.79(1.68)& 701(63.45)& 190(17.50) & 18(1.67)\\
		3D U-Net+UDA$\ddagger$ \cite{xie2020unsupervised} & 2.71(1.63)& 620(60.62)&157(15.36)& 10(0.99)\\
		3D U-Net+TS$\ddagger$ \cite{dong2019teacher} & 2.68(1.61)& 616(60.23)&154(15.06)& 9(0.91)\\
		
		3D U-Net+Fixmatch$\ddagger$ \cite{sohn2020fixmatch} & 2.73(1.63)& 623(61.03)&155(15.18)& 11(1.04)\\
		3D U-Net+Meta$\ddagger$ \cite{pham2021meta} & 2.67(1.60)& 594(58.14)&151(14.84)& 9(0.91)\\
		
		3D U-Net+ours$\ddagger$ & 2.60(1.59)& 590(57.69)&140(13.76)& 7(0.68)\\
		\cline{1-5}
		SCN \cite{Payer2019Integrating}&2.59(1.59) &627(57.58)& 160(14.67)& 10(0.98)\\
		SCN+UDA$\ddagger$ \cite{xie2020unsupervised}&2.53(1.59) &575(56.28)& 142(13.97)& 8(0.76)\\
		SCN+TS$\ddagger$ \cite{dong2019teacher}&2.54(1.61) &579(56.59)& 144(14.12)& 9(0.89)\\
		SCN+Fixmatch$\ddagger$ \cite{sohn2020fixmatch}&2.53(1.56) &576(56.38)& 141(13.89)& 8(0.76)\\
		SCN+Meta$\ddagger$ \cite{pham2021meta}&2.52(1.55) &573(56.14)& 140(13.72)& 7(0.68)\\
		SCN+ours$\ddagger$ & \textbf{2.49(1.56)}&\textbf{569(55.67)}&\textbf{126(12.31)}&\textbf{5(0.53)}\\
		\hline
	\end{tabular}}
		%\vspace{-2ex}
\end{table}

\subsubsection{\textbf{Head CBCT dataset}} It is an in-house dataset for 3D anatomical landmark localization. The dataset consists of 93 head CBCT with a size of 576 $\times$ 768 $\times$ 768, and the spacing is 0.3mm per voxel. Each labelled sample has 33 manually annotated landmarks on bone and soft tissue.
The dataset is equally distributed into 3 fixed folds: labelled training data, unlabeled data, and testing data.

\begin{table*}[h]
    %\scriptsize
	\centering
	\caption{Comparison with the different amount (proportion) of labelled data on Cephalogram dataset, Hand X-ray dataset and Head CBCT dataset. $\dag$ indicates that the baseline is applied to our semi-supervised method.}\label{tab:tab4}
	\scalebox{0.69}{
	\begin{tabular}{l c c c c| l c c c| l c c c}
		\hline
		\multicolumn{5}{c|}{Cephalogram dataset}&\multicolumn{4}{c|}{Hand X-ray dataset}&\multicolumn{3}{c}{Head CBCT dataset} \\
		\cline{1-13}
		\multirow{2}*{\tabincell{c}{Method}} & \multicolumn{4}{c|}{MRE(SD)}& \multirow{2}*{\tabincell{c}{Method}} & \multicolumn{3}{c|}{MRE(SD)}&
		\multirow{2}*{\tabincell{c}{Method}} & \multicolumn{3}{c}{MRE(SD)} \\
        \cline{2-5}
        \cline{7-9}
        \cline{11-13}
        
		%\multirow{2}*{\tabincell{c}{VGG19\_bn}} &  \multicolumn{10}{c|}{~}\\
		~ & 20 & 50 & 100 & 150& ~ & 20\% & 50\% & 100\% &~ & 20\% & 50\% & 100\% \\
		
		\hline
		U-Net& 2.09(2.84)& 1.62(1.58) & 1.48(1.49)& 1.43(1.18)&Hourglass & 0.88(0.66) & 0.78(0.66)& 0.75(0.65) &3D U-Net& 3.95(2.21)&3.11(1.84)& 2.79(1.68)\\
		U-Net$\dag$ &1.74(1.61)& 1.51(1.33) & 1.39(1.40)& 1.35(1.12)&Hourglass$\dag$ & 0.80(0.54)&0.74(0.61) &0.71(0.53)&3D U-Net$\dag$ & 3.48(1.97)& 2.86(1.71)& 2.60(1.59)\\
		AFPF-RV &1.71(1.83) & 1.46(1.11)& 1.36(1.44)& 1.27(0.99)&Modified U-Net& 0.81(0.62) & 0.71(0.56) & 0.65(0.48)&SCN &3.56(1.94)&2.82(1.70)& 2.59(1.59)\\
		AFPF-RV $\dag$ &\textbf{1.56(1.21)}& \textbf{1.36(1.01)} &\textbf{1.27(1.17)} &\textbf{1.19(0.91)}&Modified U-Net$\dag$ &\textbf{0.76(0.50)}& \textbf{0.68(0.50)} & \textbf{0.62(0.42)}&SCN$\dag$ & \textbf{3.16(1.81)}&\textbf{2.63(1.65)}& \textbf{2.49(1.56)}\\
		\hline
	\end{tabular}}
	%\vspace{-2ex}
\end{table*}

\subsection{Baselines.}

To compare our method with four other semi-supervised methods (TS \cite{dong2019teacher}, UDA \cite{xie2020unsupervised}, Fixmatch \cite{sohn2020fixmatch} and Meta \cite{pham2021meta}), we exploit the performance of semi-supervised methods applied on four backbone networks (U-Net \cite{ronneberger2015u}, Hourglass \cite{newell2016stacked}, SCN \cite{Payer2019Integrating} and AFPF-RV \cite{Chen2019cephalometric}) conducted on three datasets. Note that for SCN and AFPF-RV, their original loss function is used to perform pre-training and fine-tuning. We used the same data augmentation techniques (translation, rotation, and adding Gaussian noise) to TS and UDA for a fair comparison. All networks are optimized by Adam optimizer with default parameters on the PyTorch platform.

For \textbf{Cephalogram dataset}, two backbone networks (U-Net and AFPF-RV) are used in our framework. We adopt the same implementations of U-Net and AFPF-RV reported in Ronneberger et al. \cite{ronneberger2015u} and Chen et al. \cite{Chen2019cephalometric}, respectively. Note that the output of AFPF-RV has two offset maps and one heat map for each landmark prediction. We consider heat maps of all landmarks when processing self-training.

Two networks, including U-Net and one-stage Hourglass, are tested on \textbf{Hand X-ray dataset}. For U-Net implementation, we replace the two sequential 3 $\times$ 3 convolutional layers with a dilated block used in AFPF-RV  \cite{Chen2019cephalometric} to enlarge the receptive field and reduce the number of parameters. The input size for the modified U-Net is 512 $\times$ 512.

We adopt 3D U-Net\cite{cciccek20163d} and SCN  \cite{Payer2019Integrating} as backbone to verify the effectiveness of our method on \textbf{3D Head CBCT dataset}, where SCN is a SOTA method in 3D hand CBCT landmarks detection. The input size of both baselines is 96$\times$128$\times$128.

\subsection{Evaluation Metrics.}
We take the same evaluation metrics for comparisons with other methods on different datasets, including mean radius error (MRE), standard deviation (SD), and total number (and proportion) of outliers ($\O_{r}(in \%)$). Here $\O_{r}$ is the number of predictions larger than radius error $r$ (mm) for all testing samples. For the Cephalogram dataset and the Hand X-ray dataset, we follow the protocol of Wang et al. \cite{wang2016benchmark}, and Payer et al. \cite{Payer2019Integrating} to transfer the pixel error (number of the pixels) to the physical radius error ($r$ mm). As the spacing is 0.3mm per voxel in the 3D Head CBCT dataset, the radius error can be calculated accordingly.

\subsection{Comparison Experiments and Analysis}

As shown in Table~\ref{tab:tab1}, our method outperforms the co-training based \cite{dong2019teacher,pham2021meta} and consistency regularization based \cite{xie2020unsupervised,sohn2020fixmatch} semi-supervised method. Besides, all evaluation metrics have significant improvement when it is applied to all backbones. Therefore, we achieve new state-of-the-art performances on all three datasets. Note that the labelled data and the unlabeled data in Cephalogram and Hand X-ray datasets are from different devices, which shows that our method has superior robustness for cross-devices data.

Table \ref{tab:tab4} demonstrates the robustness of our method when applying different amounts of labelled data on three datasets. It shows that the PCA-based shape model captures the global structure even with very few labelled data. We observe that the less is the labelled data, the more significant is the performance improvement.

\subsubsection{\textbf{Cephalogram dataset}} As shown in Table ~\ref{tab:tab1}, the MRE of U-Net and the SOTA method AFPF-RV have been relatively improved by  6.3\% and 6.4\% (1.44-1.35 and 1.27-1.19), respectively, when integrated with our method. The improvement can also be seen in SD and $\O_{r}(r=2mm,2.5mm,3mm,4mm)$. Therefore, by applying our approach, the SOTA method (AFPF-RV) further improves its performance. Table~\ref{tab:tab4} demonstrates the improvement of three baselines under different amounts of labelled data. For example, the relative improvement of U-Net is 16.7\%, 7.2\%, 6.1\% and 5.9\% when using 20, 50, 100, 150 samples as labelled training data. It shows our method also performs well even with very few labelled data.

\subsubsection{\textbf{Hand X-ray dataset}} Two baselines are tested on this dataset. With 5.3\% relative improvement by our method, the modified U-Net surpasses the SOTA method in all evaluation metrics (Table~\ref{tab:tab1}). Since the labelled data and the unlabeled data are from different devices, it shows our method has superior robustness for cross-devices data.

\subsubsection{\textbf{Head CBCT dataset}} Landmark detection in 3D volume data is more challenging than 2D images. However, our method is also very effective in this type of data(Table~\ref{tab:tab1} and Table~\ref{tab:tab4}). It achieves 6.8\% and 3.8\% relative improvement for 3D U-Net and SCN by applying our method. 

\subsubsection{\textbf{Discussion}}
We discuss the advantages of our method compared with four state-of-the-art semi-supervised methods. The determination of hyper-parameters is shown. We also present the limitations of our method and future work.

\paragraph{Compared with UDA}
UDA \cite{xie2020unsupervised} apply data augmentation on unlabeled data, and then consistent regularize the predicted results of unlabeled data. It is an effective way to utilize unlabeled data to boost performance in image classification. To apply UDA in the landmark detection task, we treat the predicted landmark as the predicted class label and constraint the predicted landmarks to be identical. However, the global shape constraints are not considered in UDA, leading to some large-baized predictions are treated as pseudo labels for retraining the network. Our method filters out predictions that do not satisfy the global shape constraint, resulting in a better final performance.

\paragraph{Compared with Fixmatch}
Similar to UDA, Fixmatch \cite{sohn2020fixmatch} is also a consistency regularization-based method for semi-supervised image classification. They design some heuristics to train only the pseudo labels with high confidence. Experiments show that the global shape constraint is more effective in filtering out low-quality pseudo labels because the global shape constraint is an inherent property in the landmark detection task.

\paragraph{Compared with TS}
TS \cite{dong2019teacher} is a teacher-student style framework for semi-supervised landmark detection. The teacher network response for filtering out
unqualified pseudo labels predicted by student networks, while student networks learn from the labelled and qualified pseudo labelled samples. However, they introduce extremely more network parameters, and the training process is complex. While in our method, the PCA-based shape model is non-parameter. The affine transformation $T_{\alpha}$ is calculated by the least-squares method. Because the dimension of Shape $X$ is a small fixed number (for example, 38 in the cephalogram dataset), the time complexity is nearly O(1) to perform shape regulation on a sample. Therefore, our PCA shape model is much more efficient than the counterparts \cite{dong2019teacher,pham2021meta} that using a teacher network in filtering out low-quality pseudo labels. In addition, we find that the teacher network will fail when the labelled dataset is small (less than a hundred samples). While our PCA-based shape model has stable performance even built on a few labelled samples (20 samples). Besides, our region attention loss makes pseudo labels self-refinement and gradually closed to the unknown ground truth though iteratively self-training. Thus, by incorporating a PCA-based shape model and region attention loss, our method outperforms TS.

\paragraph{Compared with Meta}
Different from other teacher-student style frameworks, the teacher in Meta Pseudo Labels \cite{pham2021meta} is constantly adapted by the feedback of the student’s performance on the labelled dataset. However, Meta shares the identical drawback with TS in the landmark detection task. Experiments show that our method is more suitable for this task.

\paragraph{Hyper-parameters}
Our method has three hyper-parameters: 1. the number of principal components $K$ in the PCA shape model; 2. the threshold of $Z$ in abnormal detection; 3. the pre-defined Gaussian Distributions $\Delta$ in region attention loss. We use the same hyper-parameters in all experiments.

$K$ determines how much proportion of variation to preserve. There is a trade-off to set it. Generally, the shape model with a higher $K$ reserves more delicate details of global shape but is more sensitive to the outlines when performing shape adjustment. On the other hand, it will be insensitive to the abnormal landmarks with lower $K$. In our method, we choose $K$ that preserve 99.99\% of variation empirically. $Z$ is the threshold to trigger the abnormal detection process. That is to say, if a landmark's distance to its reasonable location (according to the shape model) is larger than $Z$, it probably be a noise pseudo-label and will be filtered out. So $Z$ also controls the number of pseudo labels for retraining. We set $Z$ to be 2mm as it is clinically acceptable precision. $\Delta$ determines the "step size" to search structure consistent region. To exhausts most possibilities, we set $\Delta$ to be Gaussian distributions with mean 0.01 and variance 0.005. Note that 0.01 and 0.005 are normalized by the image resolution. We find that the mean value is larger or smaller than 0.01 will reduce accuracy.

\paragraph{Limitation and future work}
Our method has some specific designs for the landmark detection task, for example, PCA-based shape model and region attention loss. Therefore, unlike other classification-oriented methods, it may not be suitable for the medical image classification task \cite{zhang2020clinically}.

Our method requires complete images and well-annotated labels to build a PCA-based shape model and pre-training the network. However, the high-quality data cannot be completely collected in practical application, especially for the high cost and high-level requirement of medical expertise in medical images. Moreover, the unlabeled medical image data are generally acquired from different devices, which are typically noise and incomplete. To this end, we may seek solutions \cite{wu2019deep,wu2020data,xia2020part} to handle the above problems in future work.

\begin{figure*}
  %\vspace*{-5ex}
  \includegraphics[width=\textwidth]{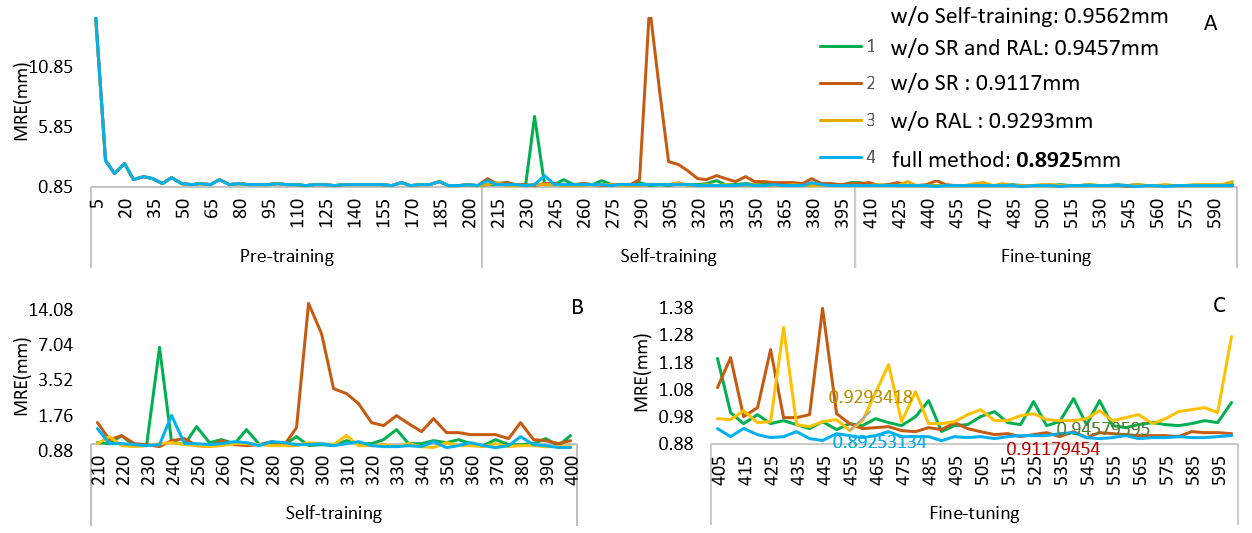}
  %\vspace*{-4ex}
  \caption{Ablation experiments on our in-house cephalogram dataset. Sub-figure A is the overview of the tested MRE(mm) along with training epochs of three stages. Sub-figure B and C are the zoom-in views of the self-training process and fine-tuning, respectively.}
  \label{fig:ablation}
  %\vspace{-2ex}
\end{figure*}

\begin{figure}
  \vspace*{-2ex}
  \includegraphics[width=0.48\textwidth]{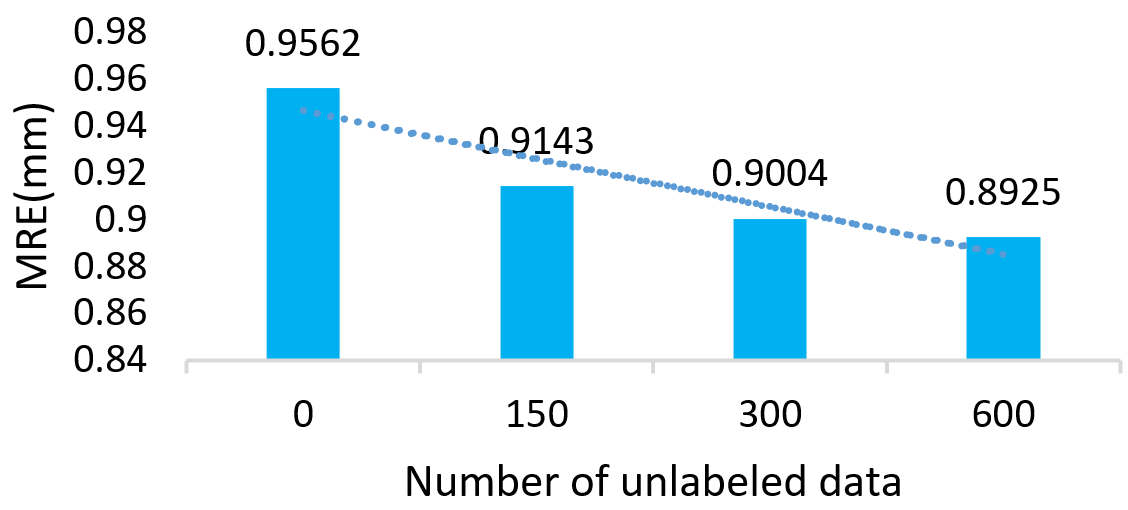}
  %\vspace*{-4ex}
  \caption{Performances for different quantities of unlabeled data.}
  \label{fig:ablation2}
  \vspace{-2ex}
\end{figure}
\subsection{Ablation study} 

This section presents ablation studies to confirm the contribution of different components in our method, including the Shape Regulation and Region Attention loss(Figure 6). We also show the performance by different quantities of unlabeled data(Figure 7). The experimental settings and analysis are described in the following.

%\vspace{-3ex}
\subsubsection{\textbf{Experimental settings}}

The ablation experiments are conducted on our large in-house dataset, which contains 900 cephalograms captured by the same device. There are 300 labelled samples with 19 manually annotated landmarks and 600 unlabeled data. The labelled data is divided into a training dataset (Train-150) and a testing dataset (Test-150), both with 150 samples. Three unlabeled datasets with different quantities of samples are randomly selected from 600 unlabeled samples, and they are named U-150, U-300, and U-600 with 150, 300, 600 samples, respectively. We conduct the following experiments to verify each component of our method: (1) remove the shape regulation and replace the region attention loss with L1 loss function (see the green line, w/o SR and RAL); (2) replace the region attention loss with L1 loss function in our full method (see the yellow line, w/o RAL); (3) our full method without Shape Regulation (see the orange line, w/o SR); (4) our full method (see the blue line, full method). We also report the performance training with only labelled data (w/o Self-training). For a fair comparison, all configurations share the same backbone network U-Net~\cite{ronneberger2015u}.

%\vspace{-3ex}
\subsubsection{\textbf{Results and Analysis}} 

Figure~\ref{fig:ablation} shows the MRE along with training epochs of three stages under different configurations. Train-150, U-600, and Test-150 are used for labelled training data, unlabeled data, and testing data, respectively. 

\paragraph{\textbf{Effect of our full method}}

Adopting the naive self-training strategy (w/o SR and  RAL) has a tiny improvement (from 0.9562mm to 0.9457mm) because some inconsistent pseudo-labelled data confuse and misinform the subsequent retraining procedure. However, benefit from the shape regulation and region attention loss to alleviate pseudo labels' inconsistency, our method achieves a noticeable accuracy improvement(from 0.9562mm to 0.8925mm).

\paragraph{\textbf{Effect of region attention loss}}

Essentially, the Region Attention loss is a soft constraint that encourages an activated region on a heatmap near the pseudo label but is not necessarily the fixed distribution centred on the pseudo label. We find this property gains the self-refinement of pseudo-labels through iteratively training. Comparing configuration 3) with 4) (yellow line and blue line) shows that Region attention loss plays a vital role in our method (0.9293mm VS 0.8925mm). 

\paragraph{\textbf{Effect of shape regulation}}
Concluded from configurations 2) and 4) (orange line and blue line), because of the shape adjustment and abnormal detection to filter inconsistent pseudo labels, the shape regulation helps stabilize the self-training process. As a result, it leads to better final performance (0.9117mm VS 0.8925mm). 

\paragraph{\textbf{Effect of unlabeled data size}}
When evaluating our full method on different quantities of unlabeled data (U-150, U-300, U-600), the related improvement are 4.4\%, 5.8\% and 6.7\%, respectively (Fig ~\ref{fig:ablation2}). It means that our method effectively learns useful features from unlabeled data. Generally, the more unlabeled data is used, the better the final performance will be.

\section{Conclusion}

We propose a novel semi-supervised approach for landmark detection in medical images. Our method computes reliable pseudo labels for unlabeled data by Shape Prior regulation. Then the Region Attention loss is used to optimize network parameters for better predicting the pseudo labels that closer its unknown ground truth. The performance is prominently improved by the self-training using unlabeled data, and our method achieves a new state-of-the-art on three medical image datasets. In addition, our method performs quite well even when the unlabeled data is from other devices and has not occurred in the labelled data. Extensive experiments have shown that our method outperforms the two semi-supervised methods on various baselines tested on 2D and 3D medical image datasets. It can also be used as a plug-and-play module integrated into other heatmap-based supervised methods to improve their performance further.


\bibliographystyle{elsarticle-num}
%\bibliographystyle{siam}

%\bibliographystyle{cas-model2-names}

\bibliography{cas-refs}

\end{document}